\documentclass[11pt]{article}

\usepackage[a4paper,left=15mm,top=20mm,right=15mm,bottom=25mm]{geometry}

\usepackage{amssymb}
\usepackage{amsthm, amsmath, amsfonts}
\usepackage{mathtools}
\usepackage{mathptmx}
\usepackage[mathscr]{euscript}

\usepackage{color}

\usepackage{hyperref}

\usepackage{caption}
\usepackage{subcaption}
\usepackage{graphicx}
\usepackage{authblk}

\usepackage[normalem]{ulem}
\makeatletter
\def\moverlay{\mathpalette\mov@rlay}
\def\mov@rlay#1#2{\leavevmode\vtop{%
		\baselineskip\z@skip \lineskiplimit-\maxdimen
		\ialign{\hfil$\m@th#1##$\hfil\cr#2\crcr}}}
\newcommand{\charfusion}[3][\mathord]{
	#1{\ifx#1\mathop\vphantom{#2}\fi
		\mathpalette\mov@rlay{#2\cr#3}
	}
	\ifx#1\mathop\expandafter\displaylimits\fi}
\makeatother

\providecommand{\keywords}[1]{\textbf{\textit{Keywords: }} #1}

\title{Fast and Automatic Object Registration for Human-Robot Collaboration in  Industrial Manufacturing}
\author[1]{Manuela Gei{\ss}}
\author[2]{Martin Baresch}
\author[1]{Georgios Chasparis}
\author[3]{Edwin Schweiger}
\author[3]{Nico Teringl}
\author[1]{Michael Zwick}

\affil[1]{Software Competence Center Hagenberg GmbH, Softwarepark 32a, 4232 Hagenberg, Austria}
\affil[2]{KEBA Group AG, Reindlstra{\ss}e 51, 4040 Linz, Austria}
\affil[3]{Danube Dynamics Embedded Solutions GmbH, Lastenstraße 38/12.OG, 4020 Linz, Austria}

\date{\ }

\begin{document}

\maketitle           
\abstract{
We present an end-to-end framework for fast retraining of object detection models in human-robot-collaboration. Our Faster R-CNN based setup covers the whole workflow of automatic image generation and labeling, model retraining on-site as well as inference on a FPGA edge device.  The intervention of a human operator reduces to providing the new object together with its label and starting the training process. Moreover, we present a new loss, the intraspread-objectosphere loss, to tackle the problem of open world recognition. Though it fails to completely solve the problem, it significantly reduces the number of false positive detections of unknown objects. 
}

\bigskip
\noindent

\sloppy

\section{Introduction}

For small-lot industrial manufacturing, reconfigurability and adaptability  becomes a major factor in the era of Industry 4.0 \cite{Mehrabi2000ReconfManufacturing}. In Smart Factories, individual production machines are equipped to an increasing degree with the ability to perform inference using machine learning models on low-resource edge devices \cite{10.1145/3469029}. In order to meet the promises of adaptability, there is an increasing demand for (re-)training models on the edge in reasonable time without the need for a cloud infrastructure \cite{Chen2019}.

State-of-the-art (deep learning) models usually operate under a closed-world assumption, i.e., all classes required for prediction are known beforehand and included in the training data set. New classes not present during training are often falsely mapped to some class close to the unknown class in feature space. However, a robot that has to react to a changing working environment is often confronted with an open world setting (open world/open set \cite{scheirer2012toward,bendale2016towards,bendale2015towards}), where new classes not present in the data during model training need to be correctly recognized with as little delay as possible. In object detection (see the use case description in Sect.~\ref{sec:usecase}), unknown objects should not be recognized as some similar object class in the training set, instead the system should initiate an incremental training step to integrate the new object class into the model. During this incremental training, automatic labeling of new training data is essential in order to avoid the time-consuming and expensive manual labeling process. In the current work, we present an end-to-end pipeline that addresses these objectives.

The remainder of this paper is structured as follows: Sect.~\ref{sec:usecase} describes the setting used by KEBA to demonstrate how to apply our pipeline in a real-world use case. The details about the used model architecture, data, and training setup are presented in Sect.~\ref{sec:model}. In Sect.~\ref{sec:AI-module}, an overview of the necessary modifications to deploy the network on the FPGA-based AI module developed by Danube Dynamics is given. Sect.~\ref{sec:autoseg} introduces our approach for automatically generating training data for new objects, especially regarding automatic segmentation. In Sect.~\ref{sec:openworld}, we address the open world challenge in object detection by developing a new loss function that achieves some improvement.% Finally, Sect.~\ref{sec:summary} summarizes our contributions, including shortcomings of our current approach, and gives an outlook into future improvements we are planning to implement.

\section{Use Case}\label{sec:usecase}
\begin{figure}[htb]
	\begin{tabular}{lcr}
		\begin{minipage}{0.6\textwidth}
			\centering
			\includegraphics[width=0.85\linewidth]{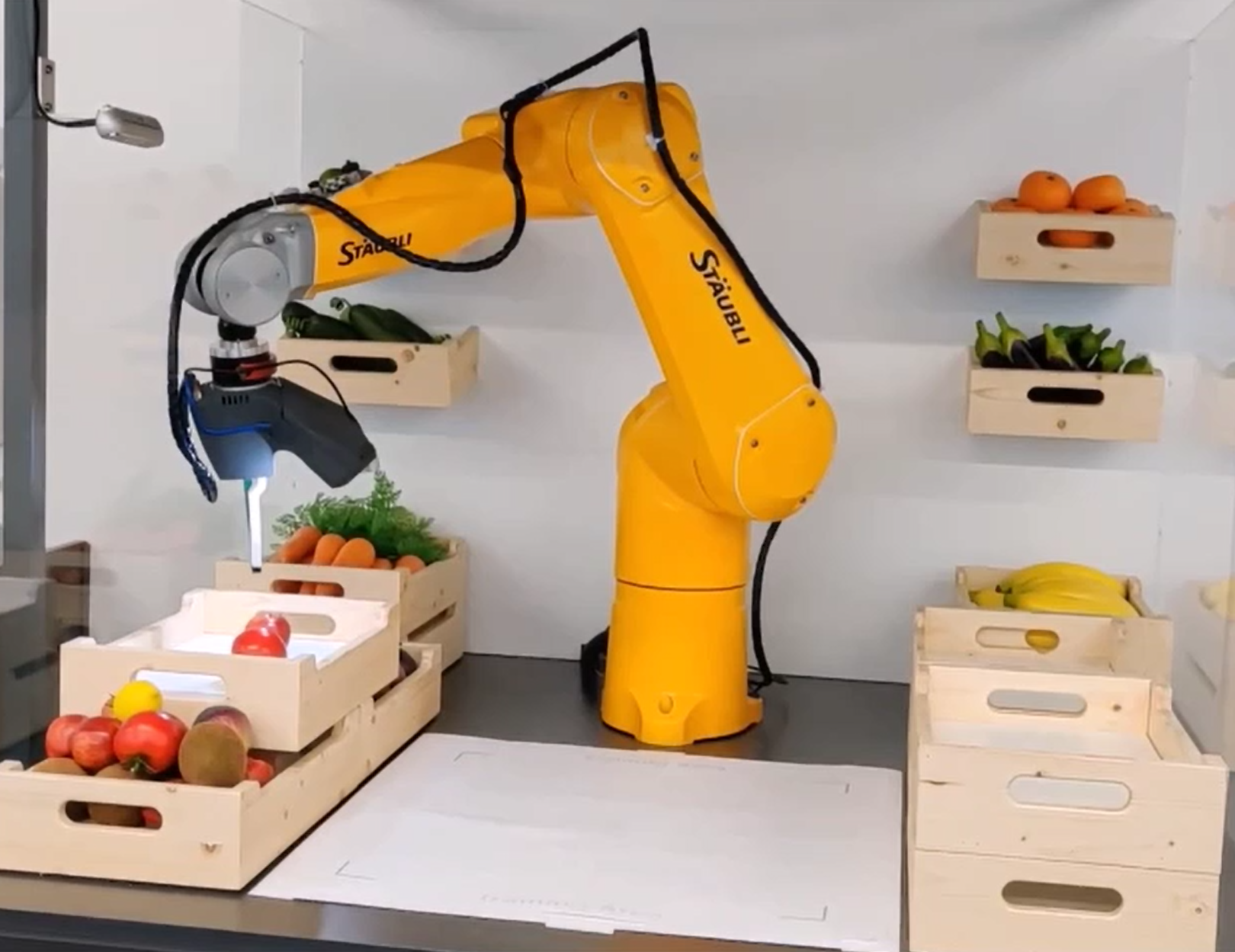}
		\end{minipage}
	 & 
		\begin{minipage}{0.3\textwidth}
			\caption{Showcase of a robot sorting different types of fruits used to demonstrate fast retraining of new classes. The left basket holds a mix of fruits which are picked up by the robot and transferred to the correct basket on the right. For generation of training images, the new object is placed in the center of the white area (Source: KEBA).}
			\label{fig:usecase}
		\end{minipage}
	\end{tabular}
\end{figure}

%\vspace{-0.6cm}
In human-robot collaboration (Cobots, \cite{matheson2019Cobot}), a robot assisting a human operator has to adapt to a changing working environment constantly. In the context of object detection, this means to be able to quickly recognize new objects present in the shared working environment, without forgetting old objects/classes \cite{ramasesh2021anatomy}.
To ensure a seamless interaction between robot and human, the robot needs to learn objects quickly (i.e.\ integrate a new class into the model), ideally in less than one minute. In such a setting, the number and variety of images for training is usually limited. Gathering images for a new object class as well as the signal to start training on this object is initiated by the operator in our setup.
For the purpose of this paper, we consider the use case of putting objects (e.g. fruits) into designated target baskets depending on the object class (see Fig.~\ref{fig:usecase}). Initially objects of different classes are located in a single basket from which the robot picks individual objects.

In addition, the model has to run on low-power/low-cost edge-devices, which puts further constraints on the model architecture as well as the data set size.

\section{Model and Training Workflow}\label{sec:model}
\subsection{Model Architecture}
We compared different state-of-the-art object detection architectures, in particular one-stage vs.\ two-state architectures. In the context of learning new objects quickly, 
we found the one-stage approach Faster R-CNN \cite{ren2015faster} to be superior to the two-stage candidates SSD \cite{liu2016ssd} and the YOLO family (YOLOv3 \cite{redmon2018yolov3}). Though the one-stage architectures needed less training time per epoch than two-stage methods, their overall training time was larger, while Faster R-CNN achieved good and robust results after a relatively small amount of training epochs. Due to the limitation in training time, we do not aim to train until the optimal point of high accuracy and low training loss but to stop training earlier without loosing too much in model performance. For this reason, we chose Faster R-CNN\footnote{\scriptsize code adapted from \url{https://github.com/yhenon/keras-frcnn/} (last pulled 02/03/2020)} with a VGG-16 backbone as the starting point for our initial framework, based on which the results of this paper have been derived. For more details on the architectures see also Fig.~\ref{fig:frcnn} and Sect.~\ref{sec:AI-module}. Note, however, that in the meanwhile new YOLO architectures have been developed at rapid pace and latest versions such as YOLOX \cite{yolox2021} are able to compete with Faster R-CNN in our application. %We are thus currently investigating and incorporating YOLO architectures into our work.

\subsection{Data and Training procedure}\label{sec:train}
In our use case, we assume that we are given a collection of base objects on which we can train the model without time constraints. As a basis for this training, we use the pretrained weights from the Keras Applications Module\footnote{\scriptsize\scriptsize\url{https://github.com/fchollet/deep-learning-models/releases} (pretrained on Imagenet)} for the VGG-16 backbone. For the subsequent steps of incrementally learning new objects, that is learning the objects that are presented to the robot by a human operator on the production site, fast training is an important issue. Moreover, it is essential that the training data of the new objects can be generated and labeled automatically, while training data for the base objects can also be labelled manually. The automatic generation of labeled training data is described in detail in Sect.~\ref{sec:autoseg}. The base training data set consists of a collection of images, each containing either a single object or multiple objects, that contains each base object approximately 150--200 times. Images of the new object, on the other hand, do only contain one instance of this new object each. We currently use $\sim$60 images of the new object with different orientations but we found that also $\sim$25 is sufficient. We currently use different collection of plastic fruits on a white background (see Fig.~\ref{fig:fruits_data} for an example). Data augmentation methods such as flipping and rotation are used during training. Given these data sets, sufficient accuracy of the base training is achieved after 30-50 minutes on an NVIDIA GTX 1080 (8GB RAM) graphics card. For the training of the new object, we merge the data set of the new and the ``old'' objects to avoid catastrophic forgetting \cite{ramasesh2021anatomy}. We observed that, although the base objects are more frequent in the data set than the new objects, this imbalance does not negatively influence the training results. For learning new objects, we tried freezing different combinations of layers (mainly in the VGG-16 backbone but also other parts) but obtained worse results, both in accuracy and training time, compared to training the whole network architecture. 

%So far, we do not use model pruning approaches for the sake of time constraints. However, with our current work on newer, improved YOLO architectures as well as the recent development on pruning methods, this aspect is an interesting topic for further research. %Moreover, we do not use few-shot methods since our training data set is already quite small such that, given the tradeoff in size and variability of the data set, we do not expect much performance increase by few-shot learning approaches in our setup. 

\section{Inference on FPGA}\label{sec:AI-module}

\begin{figure}[htb]
%	\linespread{0.7}
	\centering
	\includegraphics[width=0.8\linewidth]{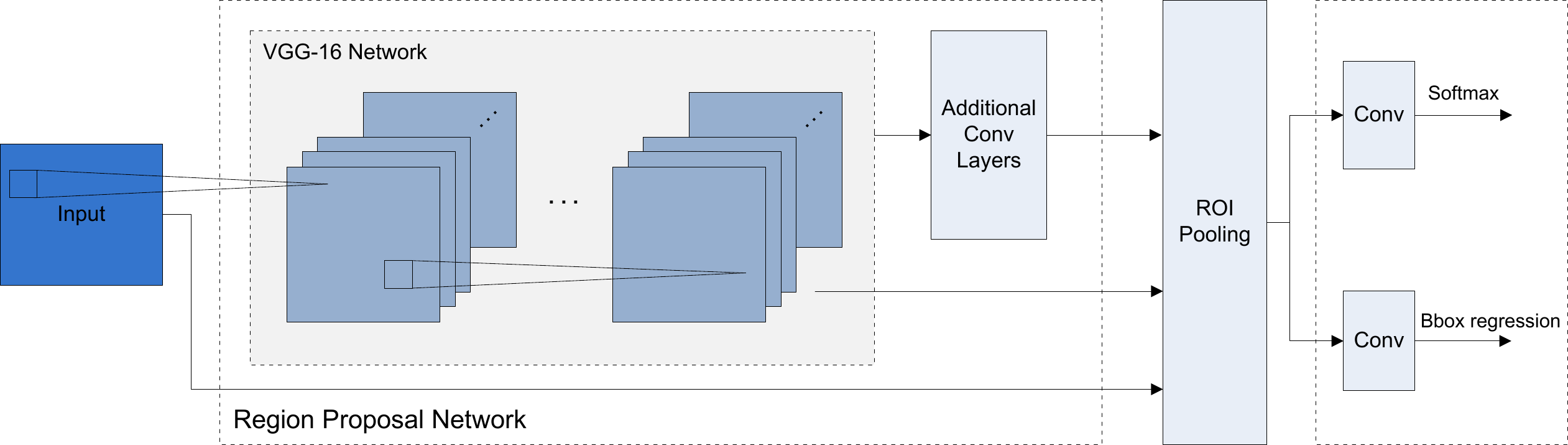}%trim=0cm 5cm 0cm 6cm, clip,
	\captionsetup{width=0.9\linewidth}
	\caption{Architecture of the used Faster R-CNN model with a VGG-16 backbone. The fully connected layers of the original Faster R-CNN architecture are replaced by convolutional layers to achieve compatibility with DNNDK. Figure adapted from \cite{zhao2016faster}.}
	\label{fig:frcnn}
\end{figure}

Using neural networks on low-resource edge devices is one vision in the context of Industry 4.0. However, current frameworks do only support inference but not training of deep neural networks on the edge. In this project, we use the Deep Neural Network Development Kit (DNNDK)\footnote{\scriptsize\scriptsize\url{https://www.xilinx.com/support/documentation/user\_guides/ug1327-dnndk-user-guide.pdf}} from Xilinx to transfer our trained Faster R-CNN model to an FPGA. However, the DNNDK framework imposes technical limitations, e.g., it requires a 4-dimensional convolutional layer as input and cannot cope with Keras' Dropout and TimeDistributed layers. For this reason, the initial Faster R-CNN architecture had to be changed, mainly by replacing fully connected layers by convolutional layers. More precisely, the network's head now consists of one single convolutional layer with Softmax activation for the classification branch and one convolutional layer with linear activation for bounding box regression (see Fig.~\ref{fig:frcnn}).

\section{Automatic generation of images and labels} \label{sec:autoseg}

For continuous and fast learning, it is crucial to automatically generate training data of the new objects in order to avoid time-consuming manual data labeling. This consists of two steps: taking pictures, and providing class labels as well as bounding box coordinates. In our setup, the new object is placed in a designated area by the human operator and the robot moves around the object and takes about 30 pictures from different angles with a camera installed on the gripper. To also capture the object's back side, the human operator then flips the object and the robot takes a second round of pictures. The class label is provided by the human operator via a user interface. For the extraction of the bounding box, we first tried different traditional image segmentation tools such as a morphological transformation (opencv: morphologyEx()\footnote{\scriptsize\url{https://docs.opencv.org/4.x/d9/d61/tutorial\_py\_morphological\_ops.html}}) 
combined with an automatic threshold method (scikit-image: local\footnote{\scriptsize\url{https://scikit-image.org/docs/dev/api/skimage.filters.html}}). 
However, despite the simplicity of our data set with objects on white background, these methods did not succeed in sufficiently eliminating noise that was caused by the lighting conditions (e.g.\ shadows). This heavily decreased the resulting accuracy of the bounding boxes.

\begin{figure}[htb]
	\begin{tabular}{lcr}
		\begin{minipage}{0.45\textwidth}
			\centering
			\includegraphics[width=0.8\linewidth]{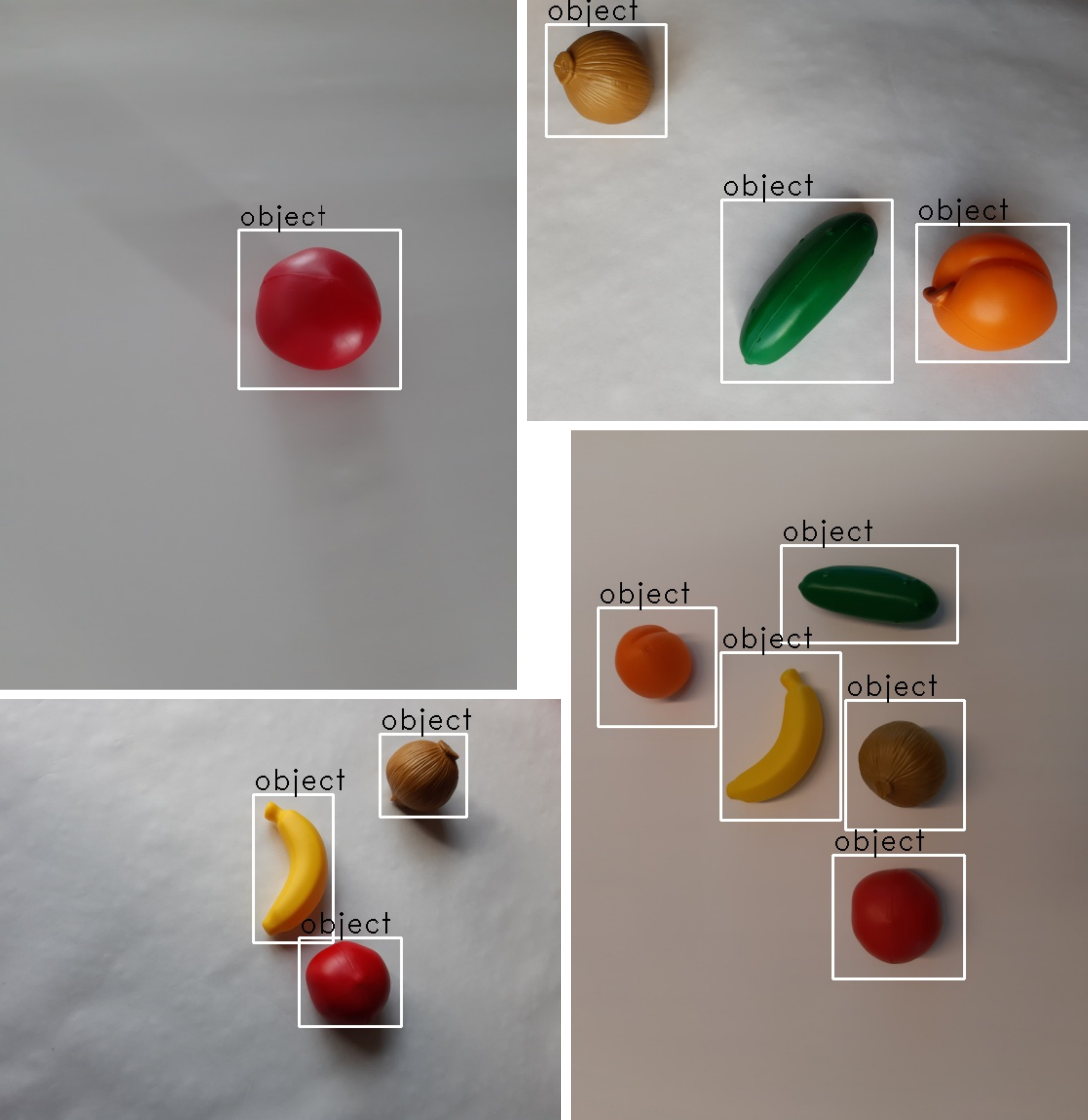}
		\end{minipage}
	 & &
		\begin{minipage}{0.4\textwidth}
			\caption[]{Examples of the data used for training the model for automatic label generation. Annotations have been generated manually. A piece of paper serves as background, however the lightning differs and often induces shadows. Images are of size 531x708 and 432x576.}
			\label{fig:fruits_data}
		\end{minipage}
	\end{tabular}
\end{figure}

\begin{figure}[htb]
	\begin{minipage}{1\linewidth}
		\subcaption[l]{\textbf{Inference results before and after post-processing}}
		%\vspace{-0.2cm}
		\centering
		\includegraphics[width=0.9\linewidth]{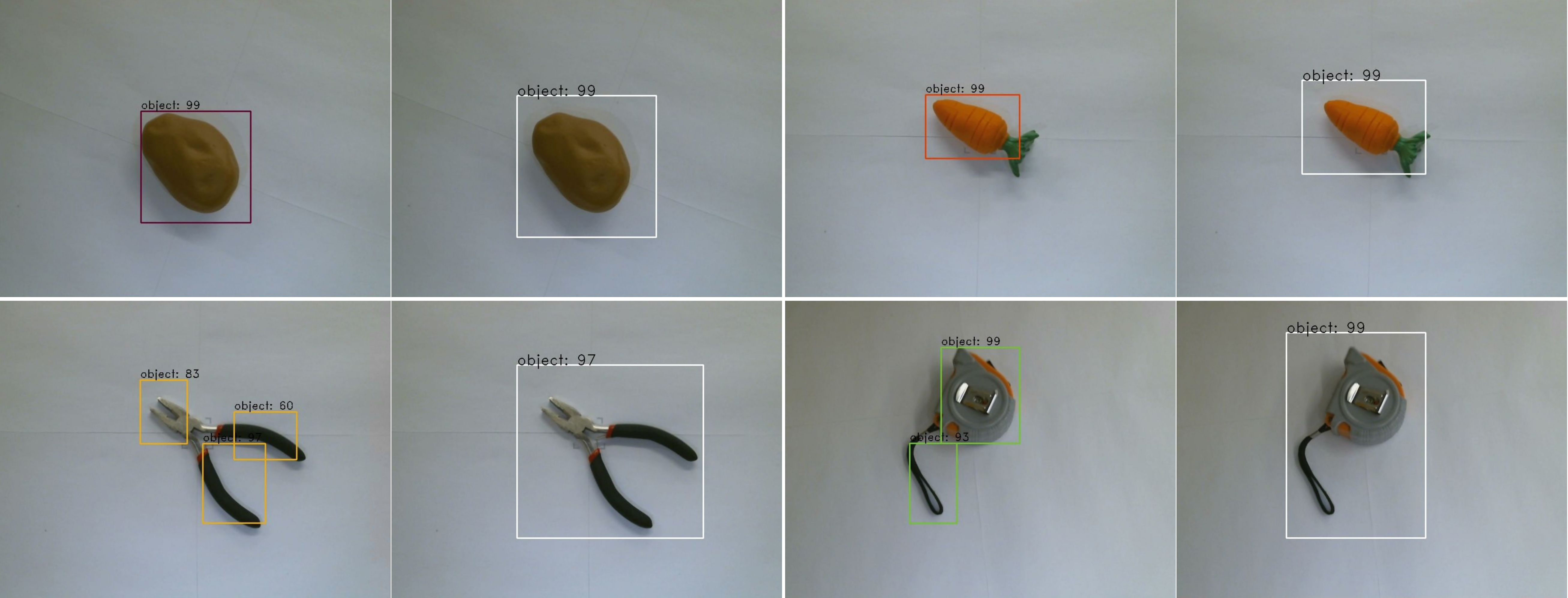}
	\end{minipage} 
	
	\begin{minipage}{1\linewidth}
		\vspace{0.5cm}
		\subcaption[l]{\textbf{Inference results on other backgrounds}}
		%\vspace{-0.1cm}
		\centering
		\includegraphics[width=0.9\linewidth]{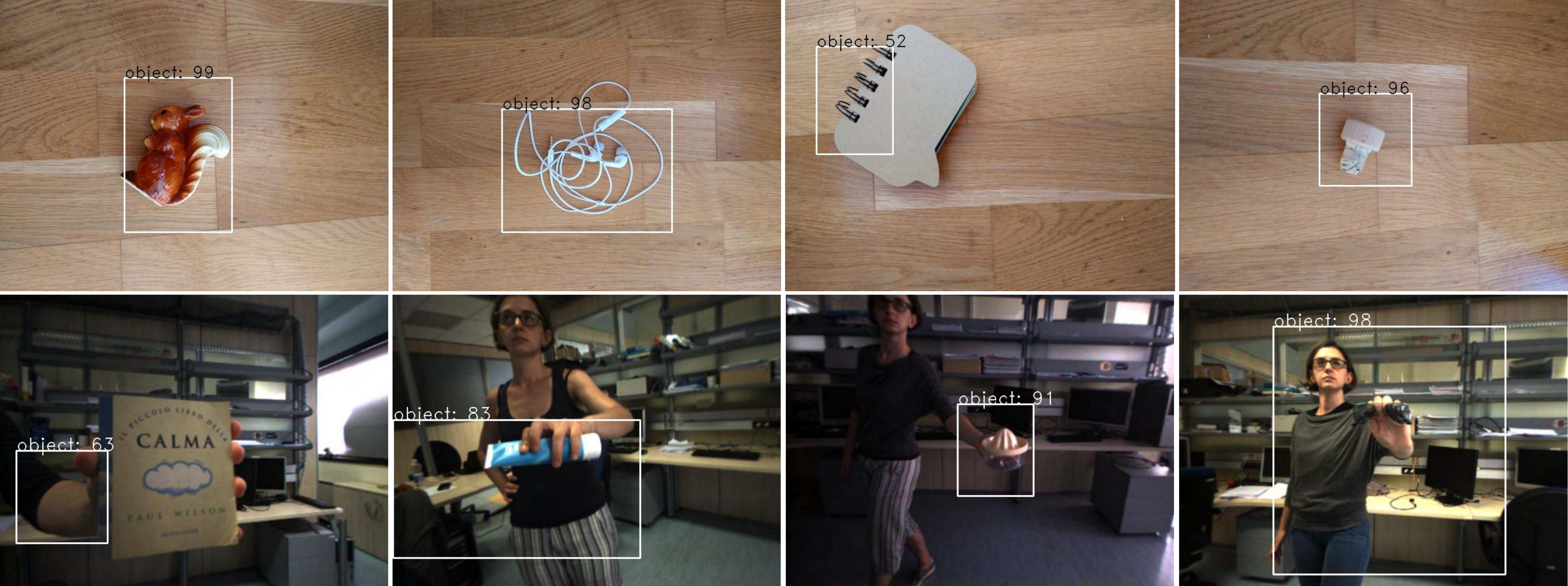}
	\end{minipage} 
	\captionsetup{width=0.9\linewidth}
	\caption{Results of the automatic generation of labeled data. The model has been trained on five fruits on a white background. (a) The raw inference results are shown on the left, the post-processed results (merging bounding boxes and adding some slack) on the right. After post-processing, all objects are well detected with high scores. (b) Inference results (after post-processing) on data sets with different backgrounds. Images in the bottom row are taken from the iCubWorld data set \cite{fanello2013icub}.
	}	
	\label{fig:results_seg}
\end{figure}

As an alternative, we trained our Faster R-CNN architecture to recognize objects by providing it manually labeled data of five different fruits that are all labeled as ``object'' (see Fig.~\ref{fig:fruits_data}). More precisely, we used the weights of the pretrained base model for object detection (see Sect.~\ref{sec:train}) and only trained the classifier part of the Faster R-CNN architecture to identify objects. The model gave good results after only a few epochs of training, the results of the final model after 200 epochs are shown in Fig.~\ref{fig:results_seg}(a) on the left. The results are equally good for objects that have not been in the training data set. As can be seen in Fig.~\ref{fig:results_seg}(a), in some cases ($\mathtt{\sim}$50\% in our setup) the bounding boxes are not ideal, that is, they either miss small parts of the object or multiple bounding boxes for different parts of the object are predicted. It is therefore beneficial to (i) merge multiple bounding boxes into one and (ii) add some additional ``slack'', i.e., increase the bounding box by a few pixel on each side. For our setup, the slightly enlarged bounding boxes do not pose a problem since the gripper takes care of fine-tuning when picking up single objects. As an interesting side effect, the model trained on white background is partly transferable to other homogeneous backgrounds (see Fig.~\ref{fig:results_seg}(b)). In summary, even though the Faster R-CNN based image labeling is slightly slower than the scikit-image based segmentation method (approx.\ 12\%) and it initially needs manual data labeling and training of the segmentation model, it has the significant advantage of being non-sensitive to reflections and therefore much more accurate.

Finally, we would like to emphasize that due to the Open World Problem (see Sect.~\ref{sec:openworld}) current state-of-the-art object detectors are not able to reliably distinguish between known and unknown objects. In our setup it is therefore only possible to automatically label images containing objects of a single class. In the specific case of our Faster R-CNN approach with the merge step during post-processing, this is even further restricted to one object per image. However, our experiments showed no significant performance difference between training images with one or multiple objects when learning a new class.

\section{Open World Challenge: The Objectosphere Loss}
\label{sec:openworld}

One of the largest limitations of current deep learning architectures is the issue of open set recognition, more precisely the incorrect detection of unknown objects (Fig.~\ref{fig:openset}, top right) that cannot be solved by simply thresholding the score functions. Inspired by \cite{dhamija2018reducing,hassen2020learning}, we developed the new \emph{intraspread-objectosphere loss} function to reduce the number of false positive detections of unknown objects.

\subsection{Methods}
Given a set of known classes $\mathcal{C}$, we write $\mathcal{X}_k$ for the set of samples belonging to a known class and $\mathcal{X}_b$ for the samples of background resp.\ unknown classes. Following the observations in \cite{dhamija2018reducing}, we aim to enlarge the variety of the background class by including some unknown objects in the training data set and handling them as background. We emphasize, however, that this can barely cover the near infinite variety of the background space.

Our new intraspread-objectosphere loss consists of two parts. The first part corresponds to the \emph{objectosphere loss}, which has been developed by \cite{dhamija2018reducing} for image classification tasks. This loss aims at concentrating background/unknown classes in the ``middle'' of the embedding space (Equ.~\ref{eq:entropic}), which corresponds to a small magnitude of the feature vector, and put some distance margin between this region and known classes (Equ.~\ref{eq:objectopshere}). The compaction of unknown classes and background is achieved by the so-called \emph{entropic open-set loss} that equally distributes the logit scores of an unknown input over all known classes as not to target one known class. This induces low feature magnitudes of these samples (see \cite{dhamija2018reducing}). This loss is given by:
\begin{equation}
    L_e(x) = \begin{cases}
		\displaystyle
		-\log S_c(x) & x\in\mathcal{X}_k\\
		-\frac{1}{|\mathcal{C}|}\sum_{c\in\mathcal{C}} \log S_c(x) & x\in \mathcal{X}_b,		
	\end{cases}
	\label{eq:entropic}
\end{equation}
where $c\in \mathcal{C}$ is the class of $x$ and  $S_c(x)=\frac{e^{l_c(x)}}{\sum_{c'\in \mathcal{C}} e^{l_{c'}(x)}}$ is the softmax score with logit scores $l_c(x)$.
To increase the distance of known classes and background, the entropic open-set loss is extended to the \emph{objectosphere loss} which puts a so-called objectosphere of radius $\xi$ around the background samples and penalizes all samples of known classes with feature magnitude inside this sphere:
\begin{equation}
    L_o(x) = L_e(x) + \lambda_o\cdot\begin{cases}
		\displaystyle
		\max(\xi - ||F(x)||, 0)^2 & x\in\mathcal{X}_k\\
		||F(x)||^2 & x\in\mathcal{X}_b,		
	\end{cases}
	\label{eq:objectopshere}
\end{equation}
where $F(x)$ (with magnitude $||F(x)||$) is the feature representation of the last network layer's input.
It has been shown in \cite{dhamija2018reducing} that $L_o$ is minimized for some input $x\in \mathcal{X}_b$ if and only if $||F(x)||=0$, which implies that the softmax scores are identical for all known classes if $x\in \mathcal{X}_b$.

For a better compaction of the known classes, we further combine the objectosphere loss with a variant $L_i$ of the \emph{intraspread loss} (see \cite{hassen2020learning}). This is tackled by using the mean feature vector of each class as the centroid of the cluster corresponding to this class. Given an input $x\in\mathcal{X}_k$ of class $c\in\mathcal{C}$, let $\mu_c$ be the mean feature vector of class $c$ taken after the previous epoch. Then, we define $L_i$ as
\begin{equation}
 L_i(x) = \sum ||\mu_c-F(x)||.  
\end{equation}
The intraspread-objectosphere loss is then given as 
\begin{equation}
L(x) = L_o(x) + \lambda_i\cdot L_i(x).
\end{equation}

\subsection{Results and Discussion}

\begin{figure}[htb]
	\centering
	\includegraphics[width=0.9\linewidth]{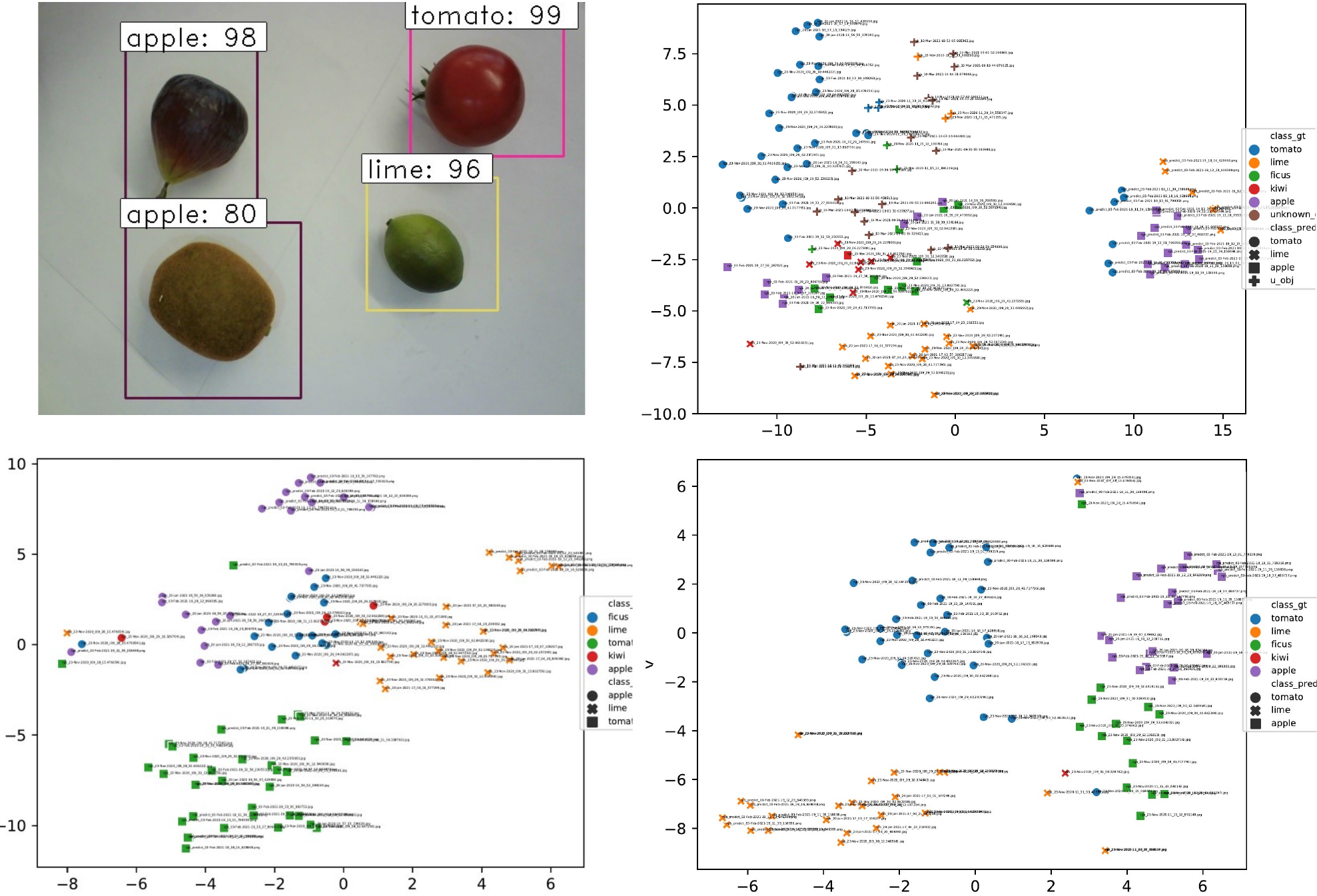}
	\captionsetup{width=0.9\linewidth}
	\caption{\textit{Top left:} Inference example after training on apple, tomato, and lime, and without the intraspread-objectosphere loss. Ficus and kiwi are mostly classified as ``apple'' with high score.
	\textit{Top right:} t-SNE embedding showing the results of the whole test data set for the experiment on the left after 30 training epochs. Colors indicate the ground truth class, symbols show the predicted class.
	\textit{Bottom left:} t-SNE embedding for the objectosphere loss after 20 epochs, $\xi=300$, $\lambda_o=10^{-4}$.
	\textit{Bottom right:} t-SNE embedding for the intraspread-objectosphere loss after 20 epochs, $\xi=300$, $\lambda_o=10^{-4}$, $\lambda_i=10^{-2}$.
	}
	\label{fig:openset}
\end{figure}

In these experiments, we focus on training the base model, again using pretrained weights from the Keras Application Module. Our training data set contains the three base classes apple, tomato, and lime %466 pics, tomato: 218, apple: 200, lime: 114
as well as 20 different unknown objects (two images each) that are added to the background class. The test data set consists of 59 images containing known objects (tomato: 35x, apple: 24x, lemon: 31x) as well as two other unknown objects on white background (ficus: 18x, kiwi: 13x) that have not been in the training data set. Using the classical Faster R-CNN loss, ficus and kiwi appear mostly as false positive detections with high score (>80\%), mainly classified as ``apple'' (see Fig.~\ref{fig:openset}, top left). Results of the whole test data set are visualized as t-SNE embedding in Fig.~\ref{fig:openset} (top right). Using the objectosphere loss reduces the number of false positives, especially for kiwi. This appears in the t-SNE embedding as less detected samples of ficus/kiwi (Fig.~\ref{fig:openset}, bottom left). It can also be seen that the classes are much better separated compared to the classical loss. The intraspread-objectsphere loss further improves this separation and further reduces the detection of false positives (Fig.~\ref{fig:openset}, bottom right). Kiwi is detected only once and ficus is detected less often and if so, then the score is rarely above 90\%. 

Hence, the intraspread-objectosphere loss clearly improves the issue of false positive detection in the context of open set recognition. However, it fails in resolving it completely.

\section{Summary}\label{sec:summary}
We presented an end-to-end framework for fast retraining of object detection models in human-robot-collaboration. It is increasingly important for machine learning models to adapt to changing environments as the integration of prediction models into the workflow of human operators (human-in-the-loop, decision support) in industrial settings advances with rapid speed. Our setup covers the whole process of retraining a model on-site, including automatic data labeling and inference on FPGA edge-devices, however, some limitations still exist: 

\begin{itemize}
    \item In order to obtain acceptable results with respect to fast retraining as well as automatic data labeling, we limit ourselves to homogeneous backgrounds. Homogeneous backgrounds are often encountered in industrial manufacturing, however, it currently limits the applicability of our approach.
    \item For our automatic labeling approach to work, new classes have to be trained using single-instance images only, i.e., each training image can contain only one object of one class.
    %New objects have to be learned one after the other, to be able to provide automatic segmentation without user intervention.
    \item One of the biggest challenges is learning in an open world, as closed world learning is still the standard for state-of-the-art object detection architectures using benchmark data sets. We introduced a new loss, the intraspread-objectosphere loss, that achieves a reduction of false positive detections but fails to completely solve the problem.
    \item With our choice of architecture (Faster R-CNN), learning on edge devices is not yet feasible.
\end{itemize}

We are currently working on porting the whole framework to a new architecture (YOLOX \cite{yolox2021}). Development of one-stage object detection architectures has been very fast paced, especially the YOLO family. By using a less resource-intensive architecture, training on low-power edge devices should become more feasible in the future. 
Due to time constraints during training, we have not investigated model pruning yet. However, with our current work on newer, improved YOLO architectures as well as the recent development on pruning methods, this aspect is an interesting topic for further research to further increase inference performance w.r.t.\ time on edge devices. Finally, the optimal training time (i.e.\ number of epochs) is heavily dependent on the use case and its data. Moreover, a new indicator for early stopping with regard to new/old objects would increase the usefulness of retraining in a collaboration environment, as standard methods like validation sets are not robust enough when training only a few epochs.

\section*{Acknowledgements}
%The research reported in this paper has been funded by the Federal Ministry for Climate Action, Environment, Energy, Mobility, Innovation and Technology (BMK), the Federal Ministry for Digital and Economic Affairs (BMDW), and the State of Upper Austria in the frame of the COMET - Competence Centers for Excellent Technologies Programme managed by Austrian Research Promotion Agency FFG.

The research reported in this paper has been funded by BMK, BMDW, and the State of Upper Austria in the frame of SCCH, part of the COMET Programme managed by FFG.

\bibliographystyle{plain} 
\bibliography{references}

\end{document}